\documentclass[5p,times]{elsarticle}
\usepackage{tabularx}
\usepackage{booktabs}
\usepackage{siunitx}
\usepackage[bookmarks=false]{hyperref}
\usepackage{subcaption}
    \hypersetup{colorlinks,
      linkcolor=blue,
      citecolor=blue,
      urlcolor=blue}
\usepackage{siunitx}

\usepackage{cleveref}
\newcolumntype{Y}{>{\raggedright\arraybackslash}X} 

\usepackage{ecrc}

\volume{00}

\firstpage{1}

\journalname{Manufacturing Letters}

\runauth{R.-S. Mei et al.}

\jid{MFGLET}

\usepackage{amssymb}

\usepackage[figuresright]{rotating}

\begin{document}

\begin{frontmatter}



\dochead{54th SME North American Manufacturing Research Conference (NAMRC 54, 2026)}

\title{Hybrid Synthetic Data Generation with Domain Randomization Enables Zero-Shot Vision-Based Part Inspection Under Extreme Class Imbalance}


\author[a]{Ruo-Syuan Mei} 
\author[a]{Sixian Jia}
\author[b]{Guangze Li\corref{cor1}}
\author[c]{Soo Yeon Lee}
\author[c]{Brian Musser}
\author[c]{William Keller}
\author[c]{Sreten Zakula}
\author[b]{Jorge Arinez}
\author[a]{Chenhui Shao\corref{cor1}}

\address[a]{Department of Mechanical Engineering, University of Michigan, Ann Arbor, MI 48109, USA}
\address[b]{Materials \textup{\&} Manufacturing Systems Research Lab, General Motors, Warren, MI 48092, USA}
\address[c]{Global Manufacturing Engineering, General Motors, Warren, MI 48092, USA}

\begin{abstract}
Machine learning, particularly deep learning, is transforming industrial quality inspection. Yet, training robust machine learning models typically requires large volumes of high-quality labeled data, which are expensive, time-consuming, and labor-intensive to obtain in manufacturing. Moreover, defective samples are intrinsically rare, leading to severe class imbalance that degrades model performance. These data constraints hinder the widespread adoption of machine learning-based quality inspection methods in real production environments. Synthetic data generation (SDG) offers a promising solution by enabling the creation of large, balanced, and fully annotated datasets in an efficient, cost-effective, and scalable manner. This paper presents a hybrid SDG framework that integrates simulation-based rendering, domain randomization, and real background compositing to enable zero-shot learning for computer vision-based industrial part inspection without manual annotation. The SDG pipeline generates 12,960 labeled images in one hour by varying part geometry, lighting, and surface properties, and then compositing synthetic parts onto real image backgrounds. A two-stage architecture utilizing a YOLOv8n backbone for object detection and MobileNetV3-small for quality classification is trained exclusively on synthetic data and evaluated on 300 real industrial parts. The proposed approach achieves an mAP@0.5 of 0.995 for detection, 96\% classification accuracy, and 90.1\% balanced accuracy. Comparative evaluation against few-shot real-data baseline approaches demonstrates significant improvement. The proposed SDG-based approach achieves 90--91\% balanced accuracy under severe class imbalance, while the baselines reach only 50\% accuracy. These results demonstrate that the proposed method enables annotation-free, scalable, and robust quality inspection for real-world manufacturing applications.
\end{abstract}

\cortext[cor1]{Corresponding authors: Guangze Li (guangze.li@gm.com), Chenhui Shao (chshao@umich.edu).}

\end{frontmatter}



\section{Introduction}
\label{sec:intro}

Visual quality inspection is critical for verifying key manufacturing indicators, including product integrity, geometric accuracy, and operational efficiency \cite{kim2018review}. Traditionally, quality inspection has relied on manual visual assessment or classical rule-based vision systems \cite{jia2025physics, islam2024deep}. Manual inspection is labor-intensive, inconsistent, and prone to human error, while rule-based systems employing template matching \cite{liu2023printing}, blob analysis \cite{he2017connected}, and edge detection \cite{canny2009computational} suffer from brittleness and poor generalization to variations in lighting, part orientation, and surface texture \cite{archana2024deep}. Although manufacturers are increasingly adopting automated inspection systems \cite{delgado2019robotics}, these limitations manifest as high false alarms and missed detection rates in practice, constraining their effectiveness in dynamic production environments.

Deep learning-based computer vision has emerged as the state-of-the-art approach for vision-based quality inspection \cite{chukwunweike2024enhancing, jia2025end}. These methods promise superior accuracy and robustness compared to manual or rule-based alternatives \cite{rauch2023semantic}. However, widespread industrial adoption faces a fundamental barrier: the data bottleneck \cite{islam2024deep, rahman2025enabling, meng2024meta}. Training robust models requires large, high-quality annotated datasets, which can be expensive, time-consuming, and difficult to acquire \cite{sundaram2023artificial, yang2019hierarchical}. Typically, thousands of images with precise bounding boxes and accurate labels \cite{kodytek2022large, villalba2019deep}. This challenge is compounded by several critical constraints inherent in manufacturing: severe class imbalance where defective samples are rare by design \cite{mehta2023greedy, jia2025end}, safety and privacy concerns in data collection and sharing \cite{MEHTA2022197, MEHTA2023687}, and computational efficiency requirements for real-time deployment on edge devices \cite{bergmann2019mvtec, liu2023survey}. Such imbalance degrades model performance where it matters most in detecting critical defects that quality systems must reliably identify.

Synthetic data generation (SDG) has emerged as a solution to address the data bottleneck, through enabling rapid, automatic creation of large-scale, balanced, annotated datasets \cite{mei2025synthetic, buggineni2024enhancing}. Recent applications demonstrate SDG's potential across diverse manufacturing domains: robot inspection \cite{zhou2024gans}, geometric defect detection \cite{mei2024deep}, semiconductor wafer analysis \cite{hu_utilizing_2024, abu_ebayyeh_improved_2022}, battery state estimation \cite{hu_state_2024}, and porosity prediction in additive manufacturing \cite{chen_dcgan-cnn_2023}. SDG methods have demonstrated capability across multiple data modalities, including 2D images \cite{werda2024generation, moonen2023cad2render, tang2023cascaded}, 3D point clouds \cite{mei2024deep}, and time-series data \cite{neunzig_enhanced_2023}. Despite these advances, direct deployment of SDG-trained models on real production data remains limited by a fundamental challenge: ``sim-to-real gap'' \cite{salvato2021crossing}. While advanced SDG methods like generative adversarial networks can produce realistic images, they often require immense computational resources \cite{brock2018large, saad2024survey}. Models trained exclusively on pristine, computer-generated images frequently fail to generalize to real manufacturing environments containing complex textures, unpredictable lighting, and cluttered backgrounds not captured in simulation \cite{zhang2023industrial, zhu2023towards}. Bridging this gap is essential for enabling SDG as a practical manufacturing solution.

This paper addresses the sim-to-real gap by combining simulation-based rendering, domain randomization, and real background compositing into a hybrid SDG framework. The approach systematically varies CAD-based part geometry, surface texture, and lighting conditions, then composites synthetic parts onto real manufacturing backgrounds to preserve geometric control while incorporating real-world appearance complexity. Applied to quality classification for automotive parts, a two-stage architecture (YOLOv8n detection and MobileNetV3-small classification) is trained on 12,960 synthetic images generated in one hour and evaluated on 300 real images under extreme class imbalance (5:1 to 11:1 pass/fail ratios). Key contributions of this work include: 
\begin{enumerate}
    \item A hybrid SDG pipeline combining simulation-based rendering, domain randomization, and real background compositing that generates 12,960 annotated images without manual labeling, enabling annotation-free model development.
    \item Zero-shot transfer achieving object detection mAP@0.5 of 0.995, classification accuracy of 96.0\%, and balanced accuracy of 90.1\% on real data, with inspection model exclusively trained on synthetic data.
    \item Robust performance under severe class imbalance, maintaining 90–91\% balanced accuracy when the pass/fail ratio reaches 5:1, while few-shot real-data baselines degrade to 50\% balanced accuracy under identical conditions.
    \item Systematic experimental evaluation comparing zero-shot SDG against few-shot real-data baselines across controlled imbalance scenarios, demonstrating an average 23.3\% improvement in balanced accuracy.
\end{enumerate}

The remainder of this paper is organized as follows. Section \ref{sec:methods} describes hybrid SDG pipeline and model training for object detection and part quality inspection. Section \ref{sec:results} presents experimental results on detection, classification, and robustness under class imbalance. Section \ref{sec:discussion} discusses model limitations, extensibility to fine-scale quality assessment, generalization to sim-to-real gap, and outlines future research direction. Finally, Section \ref{sec:conclusions} concludes the paper.

\section{Methods}
\label{sec:methods}

\begin{figure*}[h]
    \centering
    \includegraphics[width=1\linewidth]{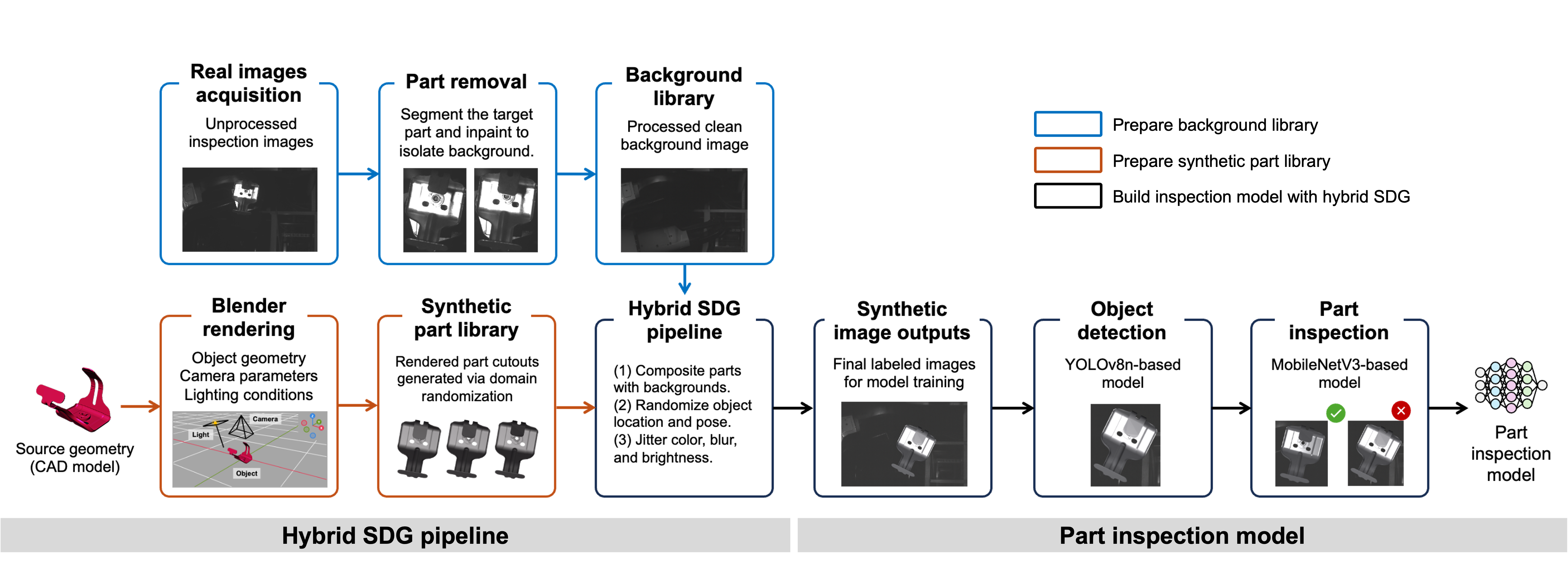}
    \caption{Proposed hybrid SDG pipeline for part quality inspection}
    \label{fig:high-level-method}
\end{figure*}

Figure~\ref{fig:high-level-method} illustrates the method overview of this study. The proposed method comprises three main components: (1) hybrid SDG pipeline to create synthetic data with automatic annotations for model training, (2) YOLOv8n model training for part detection, and (3) MobileNetV3-small training for pass/fail classification. The following subsections explain each component in details.

This study examines an industrial bracket used in automotive assembly. Bending of the top tab leads to weld defects, such as missed or edge welds, which compromise the integrity of downstream assembly operations. Example images of the bracket are used throughout this section for illustrative purposes. It is also worth noting that the proposed method is readily extensible to other, similar vision-based inspection applications.

\subsection{Hybrid SDG pipeline}
\label{sec:hybrid-sdg}

The hybrid SDG pipeline generates balanced, fully-annotated training datasets by compositing simulation-based rendered part images with real backgrounds. Unlike purely synthetic approaches that suffer from the sim-to-real gap, this hybrid strategy preserves geometric control from simulation while incorporating real-world appearance complexity to approximate production environments. As shown in Figure~\ref{fig:high-level-method}, the pipeline comprises three stages: (1) simulation-based part rendering with systematic domain randomization in Blender, (2) background extraction from real inspection station images using vision-language models, and (3) automatic composition and label generation.

\subsubsection{Part rendering with domain randomization}

\begin{figure}[h]
\centering
\includegraphics[width=0.9\linewidth]{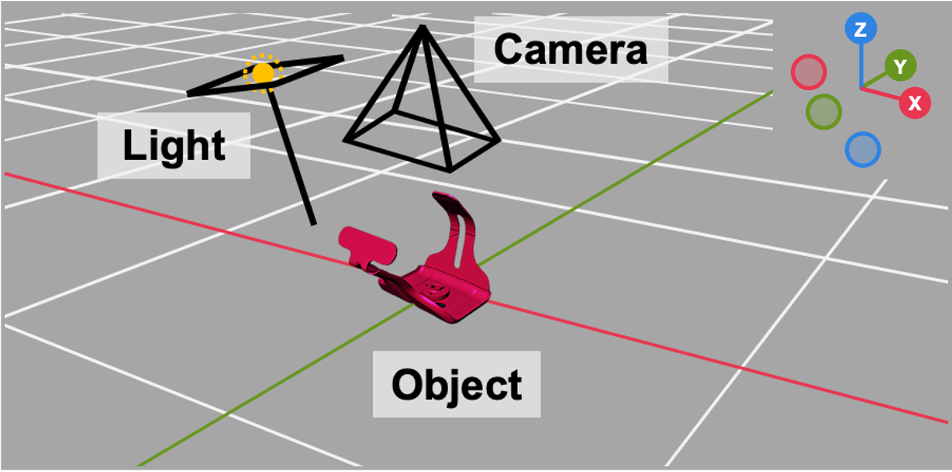}
\caption{Illustration of 3D simulation engine.}
\label{fig:blender-env}
\end{figure}

Synthetic part images are rendered in Blender 4.3.2  (Figure~\ref{fig:blender-env}) using the EEVEE real-time rendering engine. EEVEE was selected over Cycles path tracer based on computational efficiency: achieving approximately 60$\times$ faster rendering while maintaining sufficient photorealism for deep learning model training, enabling rapid dataset generation. The simulation environment comprises three components: (1) CAD-based geometric objects with adjustable mesh geometry to simulate pass and fail configurations, (2) area lights configured to mimic manufacturing floor illumination, and (3) virtual cameras calibrated to replicate inspection station optics.

The bracket CAD model was imported into Blender. Then meshes were modified to create systematic geometric variations. Tab bending angle was controlled through selective mesh editing, enabling systematic simulation of in-control configurations ($\ang{15}$, $\ang{20}$, $\ang{25}$, $\ang{30}$) and out-of-control deviations ($\ang{-5}$, $\ang{0}$, $\ang{5}$, $\ang{10}$) representative of production defects. To capture manufacturing surface finish variability, surface roughness was randomized across three levels ($r_{\textup{surface}} \in \{0.2, 0.4, 0.6\}$), while metallic properties were held constant to reflect material composition. Lighting was configured with fixed spatial position and color temperature, and light power was randomized across three levels ($P_{\textup{light}} \in \{5, 10, 15\}$ W) to simulate fill light and glare variations observed in industrial inspection environments. Virtual camera parameters (resolution, focal length, aperture, and exposure) were calibrated through comparison with real images and held constant throughout rendering to maintain consistent imaging geometry.

\begin{figure}[h]
    \centering
    \includegraphics[width=0.9\linewidth]{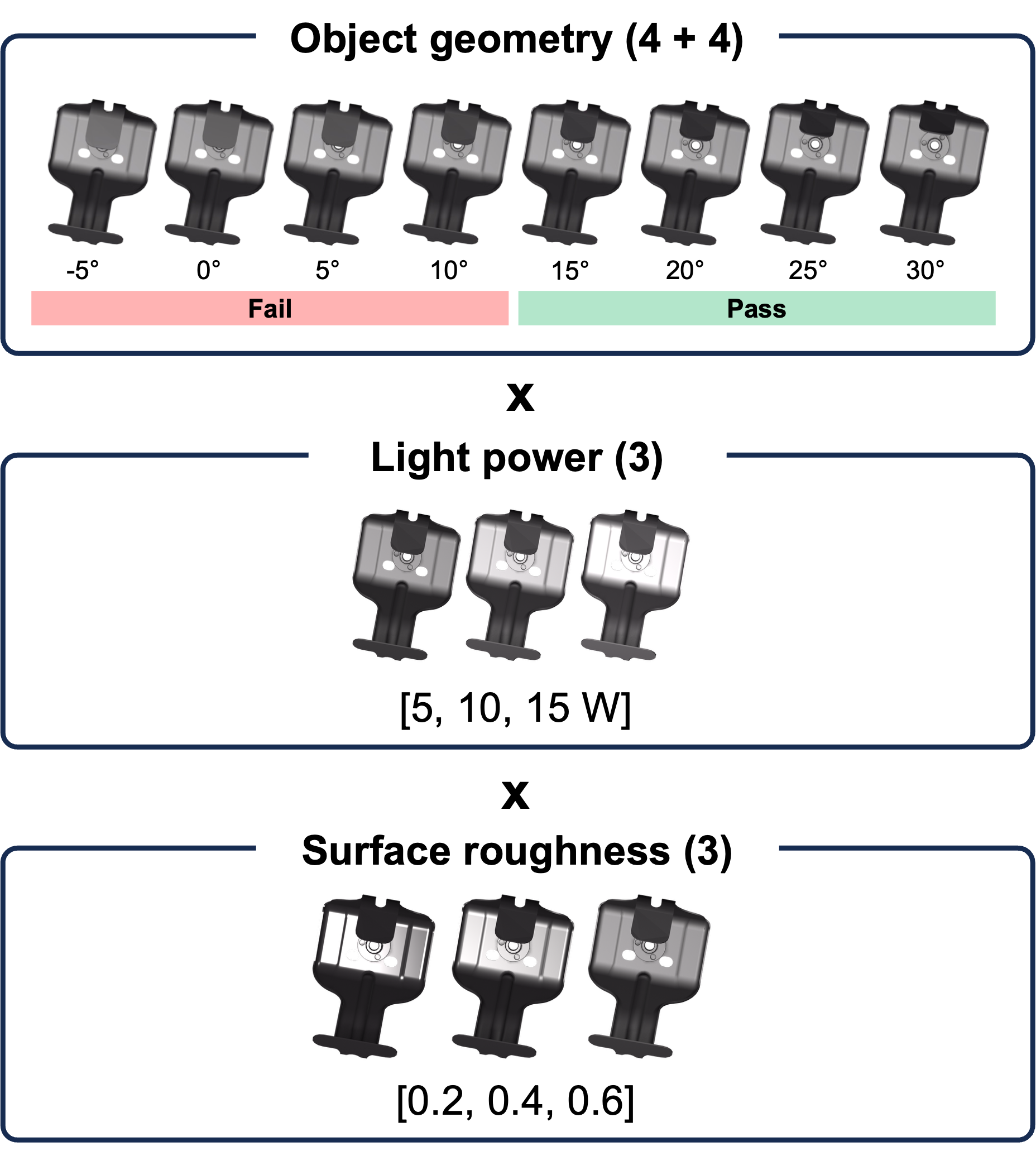}
    \caption{Domain randomization.}
    \label{fig:sdg-domain-randomization}
\end{figure}

As shown in Figure~\ref{fig:sdg-domain-randomization}, domain randomization was implemented by systematic variation of three parameters: bending angle (4 pass $\times$ 4 fail configurations), light power (3 levels), and surface roughness (3 levels). This $4 \times 3 \times 3$ factorial design per class yielded 72 unique part configurations (36 pass, 36 fail).

\subsubsection{Real background integration}

\begin{figure}[h]
    \centering
    \includegraphics[width=0.95\linewidth]{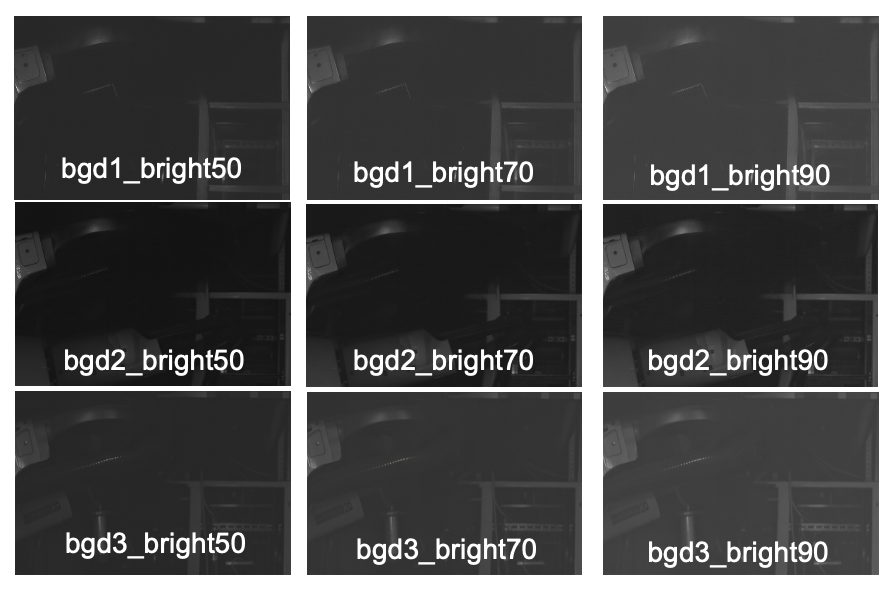}
    \caption{Background library.}
    \label{fig:bgd-lib}
\end{figure}

To bridge the sim-to-real gap, a background library was created from real images. A vision-language model was applied to remove part instances from captured images, preserving authentic environmental characteristics including lighting, shadows, and background clutter. Three representative backgrounds were selected from the inspection station environment. Background brightness was augmented across three exposure levels to simulate variations in illumination conditions, yielding 9 unique background configurations as shown in Figure~\ref{fig:bgd-lib}. For each part and background configuration, 20 images were generated using classical computer vision augmentations during model training, including rotation ($\theta_{rot} \in [$\ang{-30}$, $ \ang{+30}$]$), Gaussian blur (kernel size $\in \{1, 3, 5\}$ pixels), and brightness adjustment ([0, 50]). This approach ensures synthetic parts are composed with real environment characteristics rather than synthetic backgrounds, substantially reducing domain gap.

\subsubsection{Automated annotation generation}

\begin{figure}[h]
    \centering
    \includegraphics[width=0.95\linewidth]{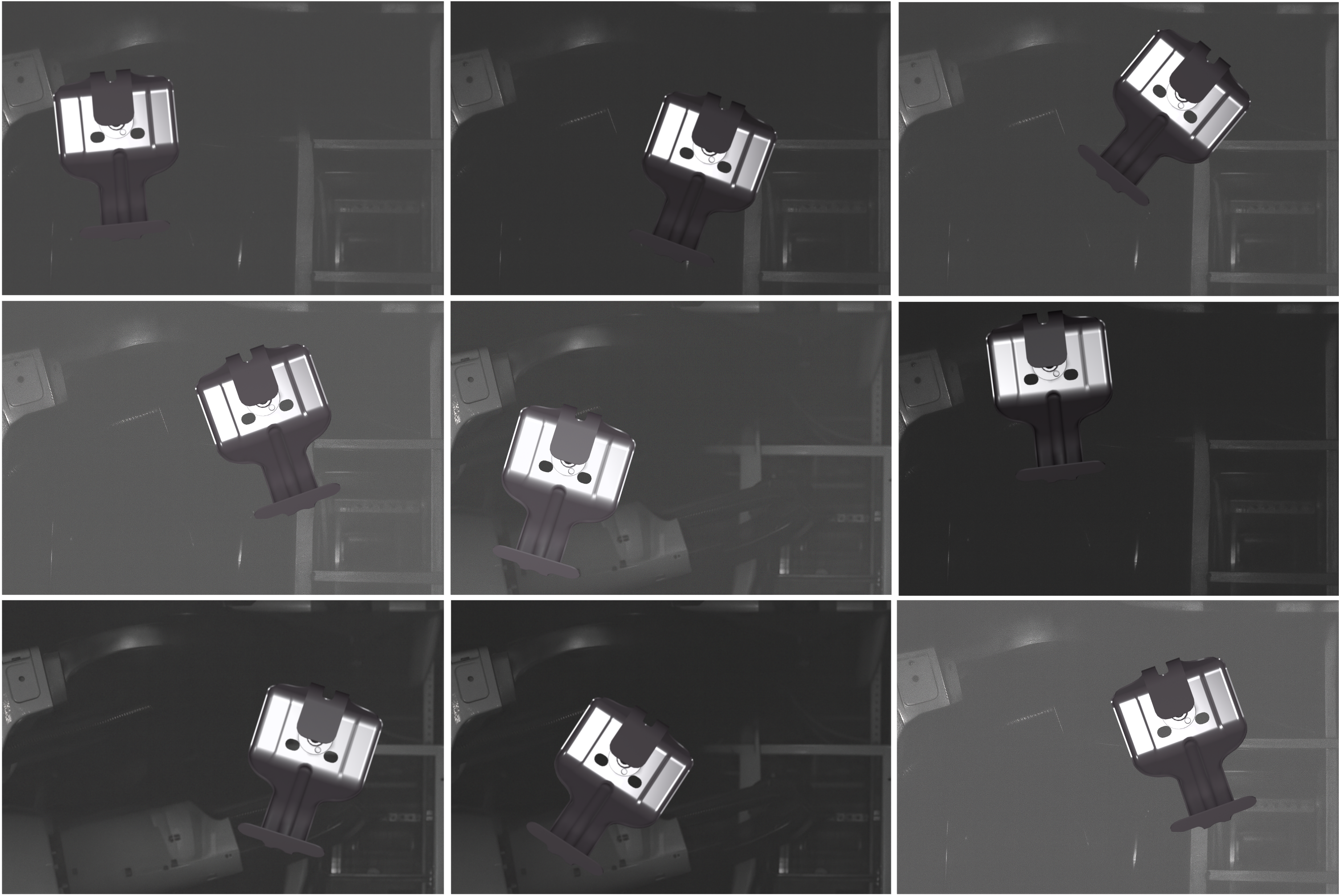}
    \caption{Examples of synthetic part images.}
    \label{fig:syn-bracket-img}
\end{figure}

Following the composition of rendered parts with real backgrounds, the pipeline automatically generates complete annotations. Bounding box coordinates for object detection are computed from the rendered 2D projections of synthetic parts. Binary class labels (pass/fail) were assigned based on the bending angle. This fully-automated annotation pipeline eliminates manual labeling, thereby enabling rapid generation of large-scale labeled datasets. We used the pipeline to generate 12,960 fully-annotated training images in under one hour on an Intel Core i5-11500H CPU (6-core, 2.9GHz) with NVIDIA T1200 Laptop GPU (4GB VRAM). The entire dataset comprises 6,480 pass images and 6,480 fail images with balanced class distribution. Some examples of synthetic part images are shown in Figure~\ref{fig:syn-bracket-img}.

\subsection{YOLOv8n for object detection}
\label{sec:yolo}
The YOLOv8n model was fine-tuned for part detection using transfer learning from COCO-pretrained weights. Training was conducted on synthetic images generated through the hybrid SDG pipeline The model was trained on an NVIDIA T1200 Laptop GPU, with input resolution standardized to 640×640 pixels. Bounding box annotations were generated automatically during SDG by recording the spatial coordinates where part objects were overlaid onto background images through classical augmentation transformations. This approach eliminated manual annotation requirements while ensuring pixel-perfect localization ground truth. The lightweight YOLOv8n architecture (3.2M parameters) was selected to balance detection accuracy with real-time inference requirements for deployment on edge devices. Transfer learning from COCO weights enabled rapid convergence despite the limited training epochs, leveraging pre-learned low-level features for part localization while adapting high-level layers to the manufacturing inspection task.

\subsection{MobileNetV3-small for quality inspection}
\label{sec:mobilenet}
The MobileNetV3-small model was fine-tuned for binary pass/fail classification using transfer learning from ImageNet-1k pretrained weights. Training was conducted on ground truth cropped part regions extracted from the same synthetic images used for YOLOv8n training, ensuring consistent data distribution while eliminating detection errors during classifier training. Input images were resized to 224×224 pixels, randomly flipped horizontally, and normalized using ImageNet statistics (mean=[0.485, 0.456, 0.406], std=[0.229, 0.224, 0.225]). The model was trained with a batch size of 4 on an NVIDIA T1200 Laptop GPU using the AdamW optimizer. Additional data augmentation was applied during training through ColorJitter transformations (brightness=0.1, contrast=0.1) to enhance robustness to lighting variations in manufacturing environments. The final classification layer was modified from 1000 ImageNet classes to 2 output classes (pass/fail) while retaining all pretrained feature extraction layers. MobileNetV3-small was selected for its compact architecture (2.5M parameters) and computational efficiency, enabling real-time classification on resource-constrained edge devices following YOLOv8n part detection.

\section{Results}
\label{sec:results}

\subsection{Part detection performance}
\label{sec:part-detection}

Table \ref{tab:yolo-finetune} presents object detection performance comparing the COCO-pretrained baseline with the proposed hybrid SDG fine-tuning approach. The pretrained YOLOv8n model without domain-specific adaptation achieves $mAP@0.5$ of only 0.089, demonstrating substantial domain gap between general object detection capabilities and manufacturing-specific inspection requirements. Despite achieving high recall (0.933), the extremely low precision (0.060) indicates the model generates excessive false detections when applied directly to factory floor images, resulting in 1 correct detection per 16 predictions.

Fine-tuning YOLOv8n using the hybrid SDG pipeline achieves mAP@0.5 of 0.995 and $mAP@0.5:0.95$ of 0.986, representing $11.2\times$ and $14.7\times$ improvements over the pretrained baseline, respectively. Precision improves from 0.060 to 0.996 ($16.6\times$ gain) while maintaining perfect recall (1.000), effectively eliminating false detections while preserving complete part coverage. This performance demonstrates effective sim-to-real transfer enabled by the hybrid approach, bridging the domain gap between synthetic training data and real-world manufacturing environments. The fine-tuned model maintains computational efficiency with inference time of 7.6 ms per image on the NVIDIA T1200 GPU, enabling real-time inspection for deployment on edge devices.

\begin{table}[h]
\centering
\caption{Object detection performance: COCO-pretrained baseline vs. hybrid SDG fine-tuned YOLOv8n}
\label{tab:yolo-finetune}
\small
\begin{tabular}{lcc}
\toprule
\textbf{Metric} & \textbf{Pretrained} & \textbf{Fine-tuned} \\
\midrule
$mAP@0.5$ & 0.089 & 0.995 \\
$mAP@0.5:0.95$ & 0.067 & 0.986 \\
Precision & 0.060 & 0.996 \\
Recall & 0.933 & 1.000 \\
\bottomrule
\end{tabular}
\end{table}

\vspace{-4pt}

\subsection{Classification performance on real data}
\label{sec:sdg-performance}

\begin{figure}[h]
    \centering
    \includegraphics[width=0.75\linewidth]{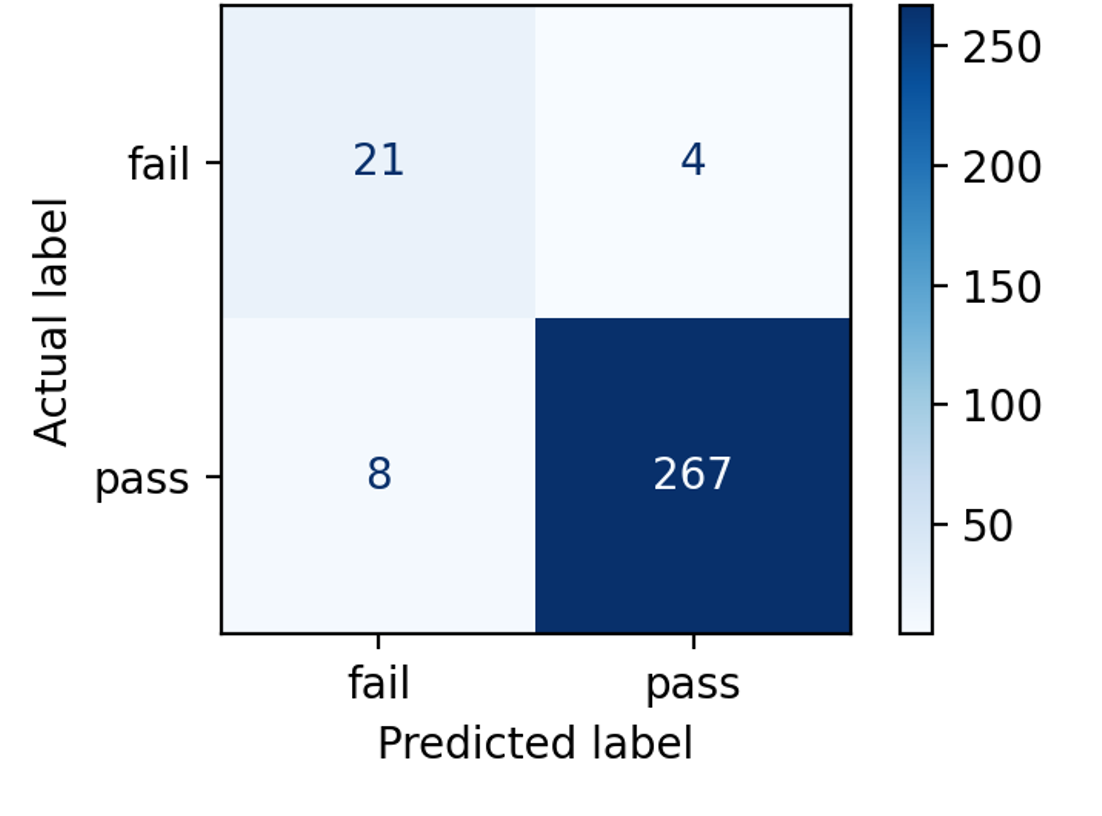}
    \caption{Confusion matrix of the SDG model evaluated on the complete dataset of 300 images. The model achieves a classification accuracy of 96\%, correctly identifying 267 images as pass and 21 images as fail. Misclassifications consist of 8 Type I errors (false alarm: unnecessary rejection of good items) and 4 Type II errors (missed detection: allowing defective items through), totaling 12 misclassified instances.}
    \label{fig:cm_sdg_model}
\end{figure}

Table~\ref{tab:sdg-model-performance} and Figure~\ref{fig:cm_sdg_model} present the classification performance of the proposed approach. Our SDG model achieves an overall accuracy of 96\% on the 300 real images. The model demonstrates consistent performance across both classes. The pass class achieves precision (0.985) and recall (0.971), correctly classifying 267 of 275 pass samples. The fail class similarly exhibits performance with precision (0.724) and recall (0.840), successfully identifying 21 of 25 defects. These results demonstrate that SDG effectively learns discriminative features from synthetic data and generalizes to real samples.

The model's strong recall on the fail class (0.840) is particularly notable for quality control applications, where detecting defects is pivotal. Only 4 fail instances are misclassified as pass, while 8 pass instances are incorrectly rejected. A manageable error distribution that confirms the model's capability to maintain high detection sensitivity. The weighted F1-score of 0.961 and weighted recall of 0.960 further validate that SDG achieves excellent performance despite the inherent class imbalance in the dataset.

\begin{table}[h]
\centering
\caption{Classification performance on 300 real images. The SDG model achieves 96.0\% accuracy and 90.1\% balanced accuracy under 11:1 class imbalance (275 pass, 25 fail samples).}
\label{tab:sdg-model-performance}
\small
\begin{tabular}{lcccc}
\toprule
\textbf{Class} & \textbf{Precision} & \textbf{Recall} & \textbf{F1-Score} & \textbf{\# Images} \\
\midrule
Pass & 0.985 & 0.971 & 0.978 & 275 \\
Fail & 0.724 & 0.840 & 0.778 & 25 \\
\midrule
Macro Avg & 0.855 & 0.905 & 0.878 & 300 \\
Weighted Avg & 0.963 & 0.960 & 0.961 & 300 \\
\midrule
Accuracy & -- & -- & 0.960 & 300 \\
\bottomrule
\end{tabular}
\end{table}

\subsection{Evaluation under balanced few-shot scenarios}
\label{sec:balanced-few-shot-baseline}

To evaluate SDG model generalization, few-shot learning baselines are established by fine-tuning MobileNetV3-small on small quantities of real data. Equal numbers of pass and fail images are randomly selected for training the few-shot baseline (FS-Real), with remaining images reserved for evaluation of both models. Experiments span 2, 4, 6, 8, and 10 shots per class, with five repetitions each. Balanced accuracy is used as the primary metric throughout this evaluation to account for class imbalance in the validation set and enable fair comparison under varying data distributions.

\begin{figure}[h]
    \centering
    \includegraphics[width=0.85\linewidth]{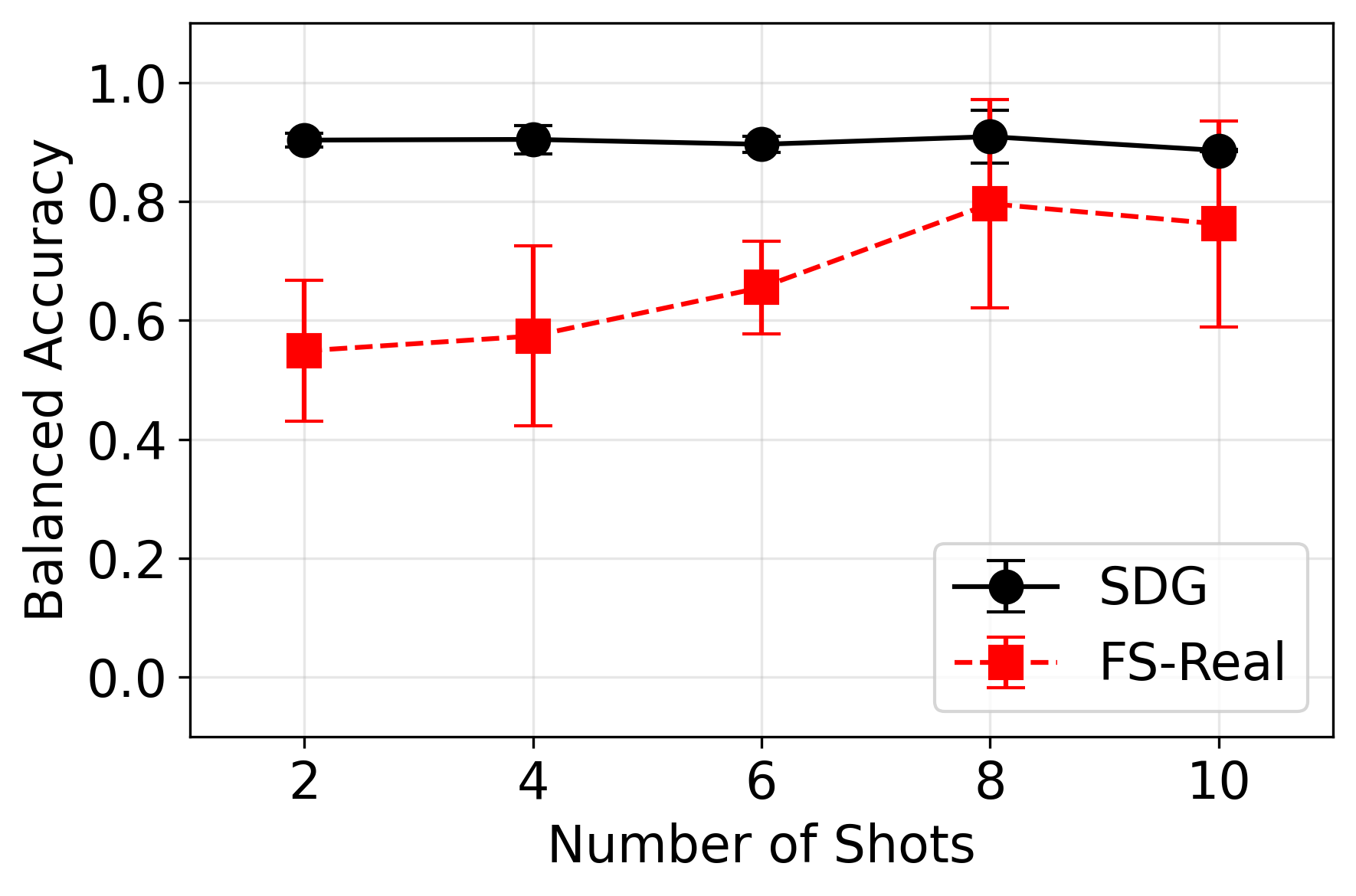}
    \caption{Balanced accuracy comparison between SDG and FS-Real across varying numbers of training shots. Error bars represent half-range values from five experimental repetitions.}
    \label{fig:bacc_SDG_FS-Real}
\end{figure}

Table~\ref{tab:bacc-sdg-fs} and Figure~\ref{fig:bacc_SDG_FS-Real} compare the proposed method with FS-Real with varied availability of real images. It is demonstrated that the SDG model consistently outperforms FS-Real across all shot configurations. The SDG model achieves average balanced accuracy of 0.900 compared to 0.667 for FS-Real, representing a 0.233 improvement. The SDG model maintains stable performance (0.886–0.909) with low half-range (0.002–0.044), while FS-Real exhibits both lower performance and higher half-range (0.078–0.175). The performance gap is largest at low shot counts: with only 2 training samples per class, SDG achieves 0.903 balanced accuracy while FS-Real achieves 0.549, demonstrating the practical advantage of synthetic data when real labeled data is scarce.

\begin{table}[h]
\centering
\caption{Balanced accuracy comparison of SDG and FS-Real under balanced pass/fail training shots}
\label{tab:bacc-sdg-fs}
\small
\begin{tabular}{lcc}
\toprule
\textbf{\# Shots} & \textbf{FS-Real} & \textbf{SDG (Ours)} \\
\midrule
2  & $0.549 \pm 0.118$ & $0.903 \pm 0.012$ \\
4  & $0.574 \pm 0.151$ & $0.904 \pm 0.024$ \\
6  & $0.655 \pm 0.078$ & $0.896 \pm 0.013$ \\
8  & $0.796 \pm 0.175$ & $0.909 \pm 0.044$ \\
10 & $0.762 \pm 0.173$ & $0.886 \pm 0.002$ \\
\midrule
Average & 0.667 ± 0.139 & 0.900 ± 0.019 \\
\bottomrule
\end{tabular}
\end{table}

\begin{figure}[h]
    \centering
    \includegraphics[width=1\linewidth]{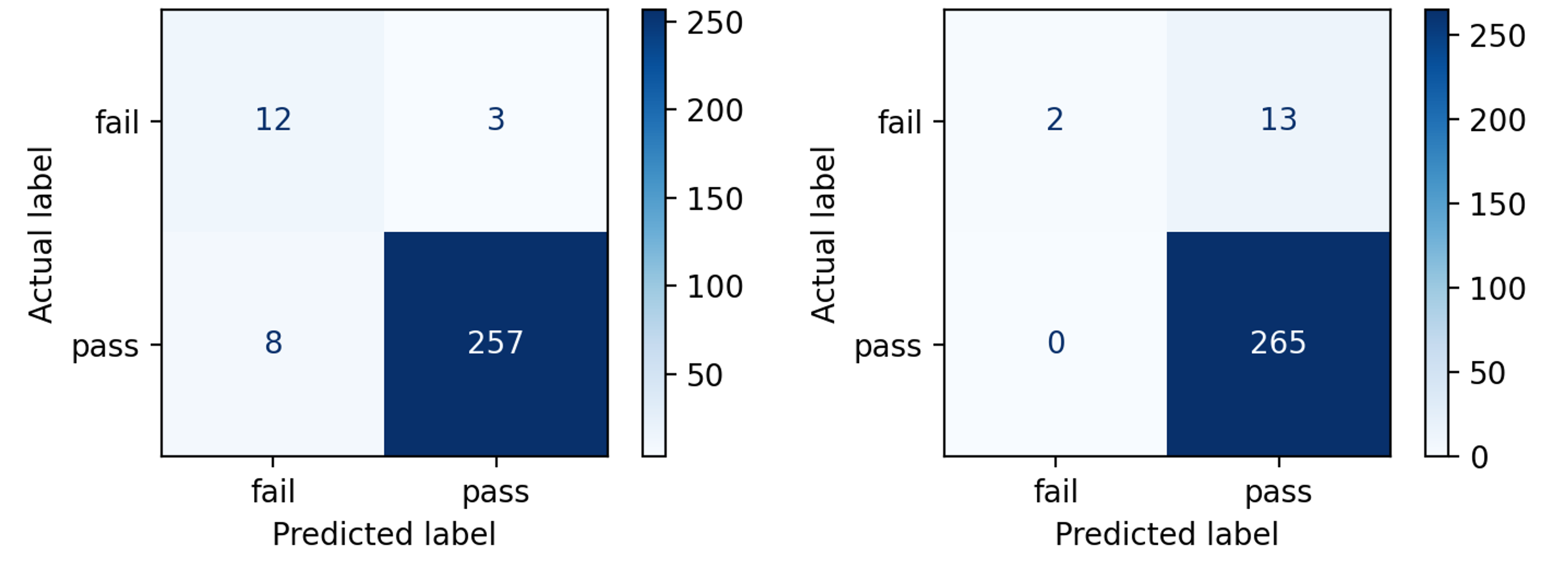}
    \caption{Confusion matrices for SDG (left) and FS-Real (right) where 10 pass images and 10 fail images were used for fine-tuning the FS-Real model.}
    \label{fig:cm_ps10_fs10_seed43}
\end{figure}

Figure~\ref{fig:cm_ps10_fs10_seed43} presents the confusion matrices comparing the SDG and FS-Real models on a held-out validation set of 280 images, with 10 pass and 10 fail examples reserved for FS-Real model training. The SDG model achieves 269 correct classifications out of 280 samples (96.1\% accuracy), correctly identifying 12 fail parts and 257 pass parts. In contrast, the FS-Real model correctly identifies only 2 out of 15 fail samples (13.3\% detection rate for fail class). This performance difference is evident in the Type II error rates: the SDG model misclassifies 3 fail parts as pass, whereas the FS-Real model misclassifies 13 fail parts as pass, representing a 4.3× increase in defects being incorrectly labeled as conforming. The high Type II error rate in FS-Real results in a higher proportion of non-conforming products being classified as conforming, which presents increased risk in quality assurance processes.

\subsection{Evaluation under imbalanced few-shot scenarios}
\label{sec:imbalanced-few-shot-baseline}

To evaluate SDG performance in realistic production settings with imbalanced data distributions, we compare it against a FS-Real baseline across scenarios where pass examples exceed fail examples. We systematically vary pass and fail shot counts across 2, 4, 6, 8, and 10 shots, maintaining pass shots $\geq$ fail shots to reflect real-world deployment conditions.

\begin{figure}[h]
    \centering
    \includegraphics[width=1\linewidth]{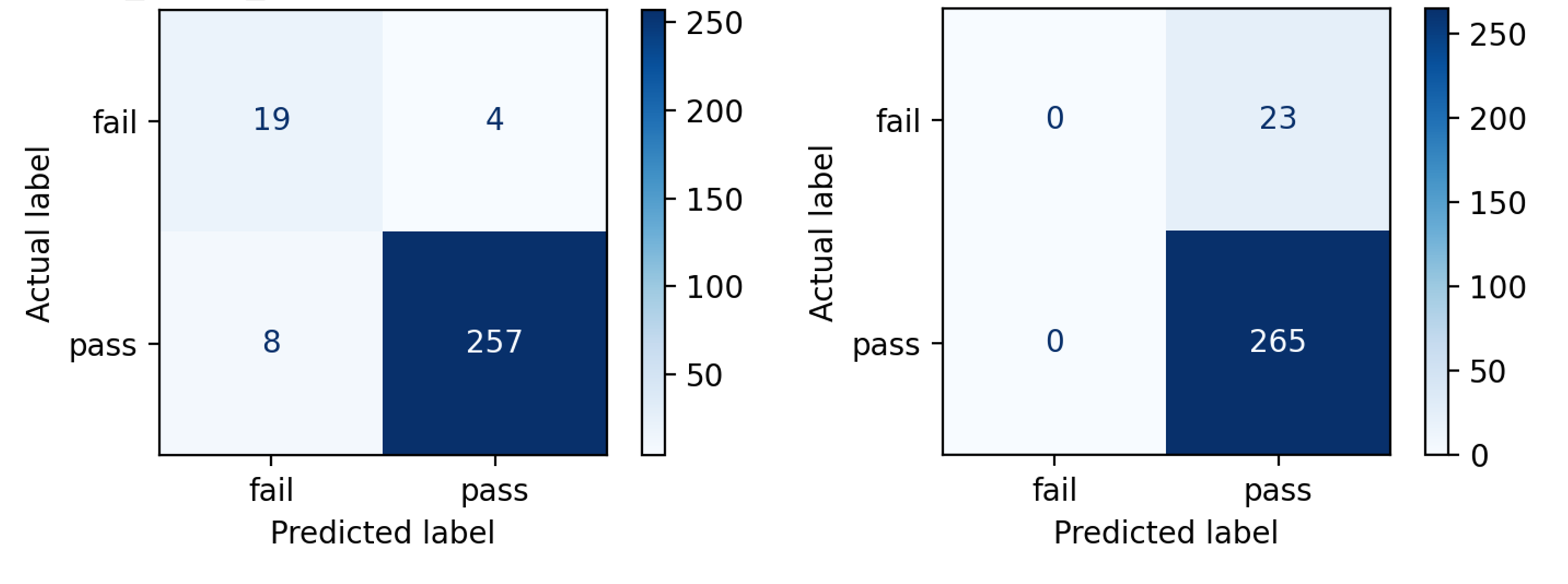}
    \caption{Confusion matrices for SDG (left) and FS-Real (right) where 10 pass images and 2 fail images were used for fine-tuning the FS-Real model.}
    \label{fig:cm_ps10_fs2_seed43}
\end{figure}

Figure~\ref{fig:cm_ps10_fs2_seed43} compares confusion matrices for the extreme imbalance case (10 pass shots, 2 fail shots) on 288 held-out samples. The SDG model classifies 276 samples correctly (95.8\% accuracy), identifying 19 fail parts and 257 pass parts correctly. The FS-Real model classifies all 288 samples as pass, with 0 fail samples identified out of 23 total fail samples (0\% detection rate). The Type II error (fail predicted as pass) differs substantially: SDG shows 4 misclassifications while FS-Real shows 23, representing a 5.8× difference. This disparity reflects the SDG model's capability to detect non-conforming products under extreme data imbalance. The elevated Type II error rate in FS-Real results in a higher proportion of non-conforming products being classified as conforming, which introduces risk in quality assurance operations.

\begin{figure*}[t]
    \centering
    \begin{subfigure}[b]{0.33\linewidth}
        \centering
        \includegraphics[width=\linewidth]{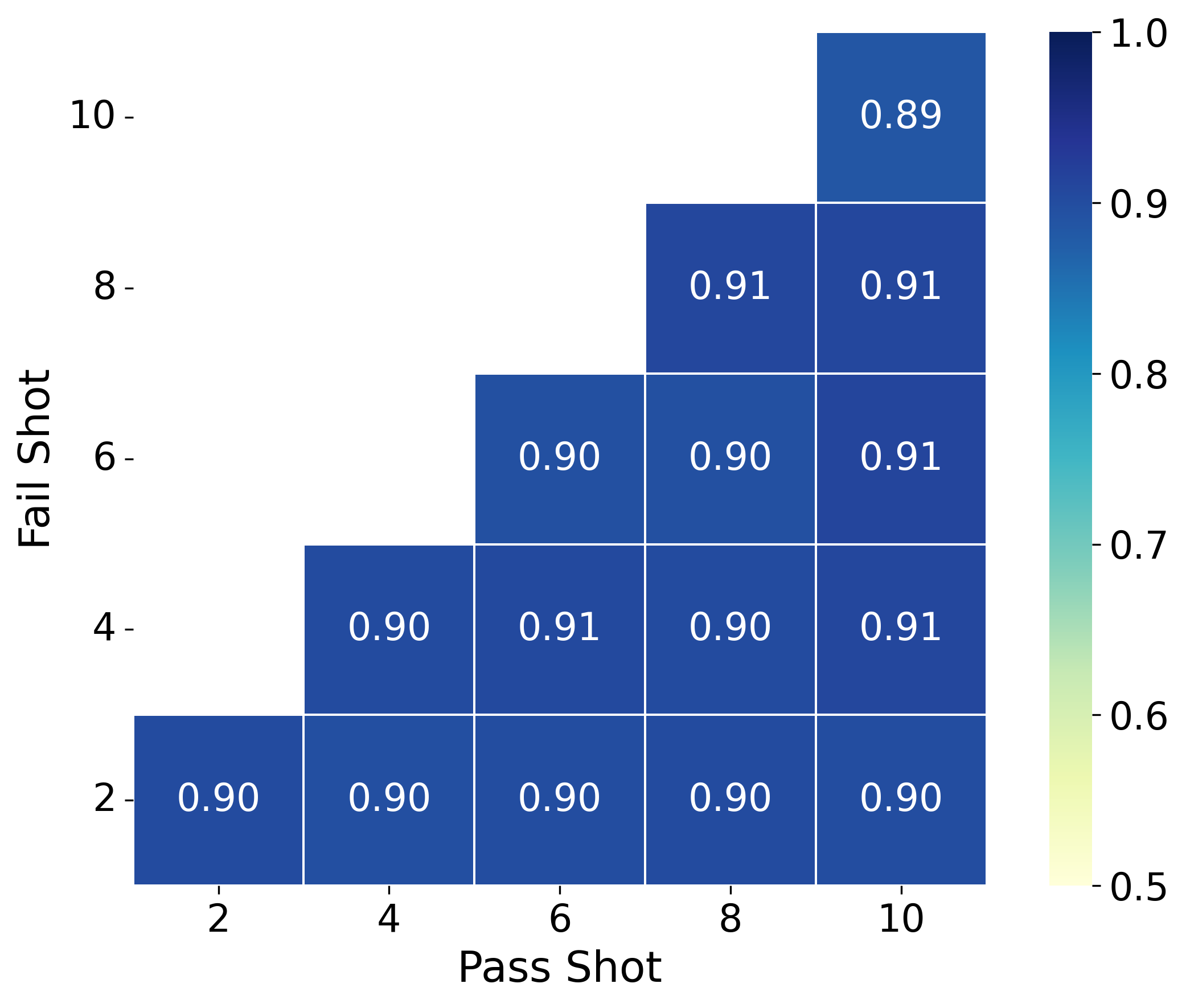}
        \caption{SDG}
        \label{fig:heatmap_imbalanced_sdg_model}
    \end{subfigure}
    \hfill
    \begin{subfigure}[b]{0.33\linewidth}
        \centering
        \includegraphics[width=\linewidth]{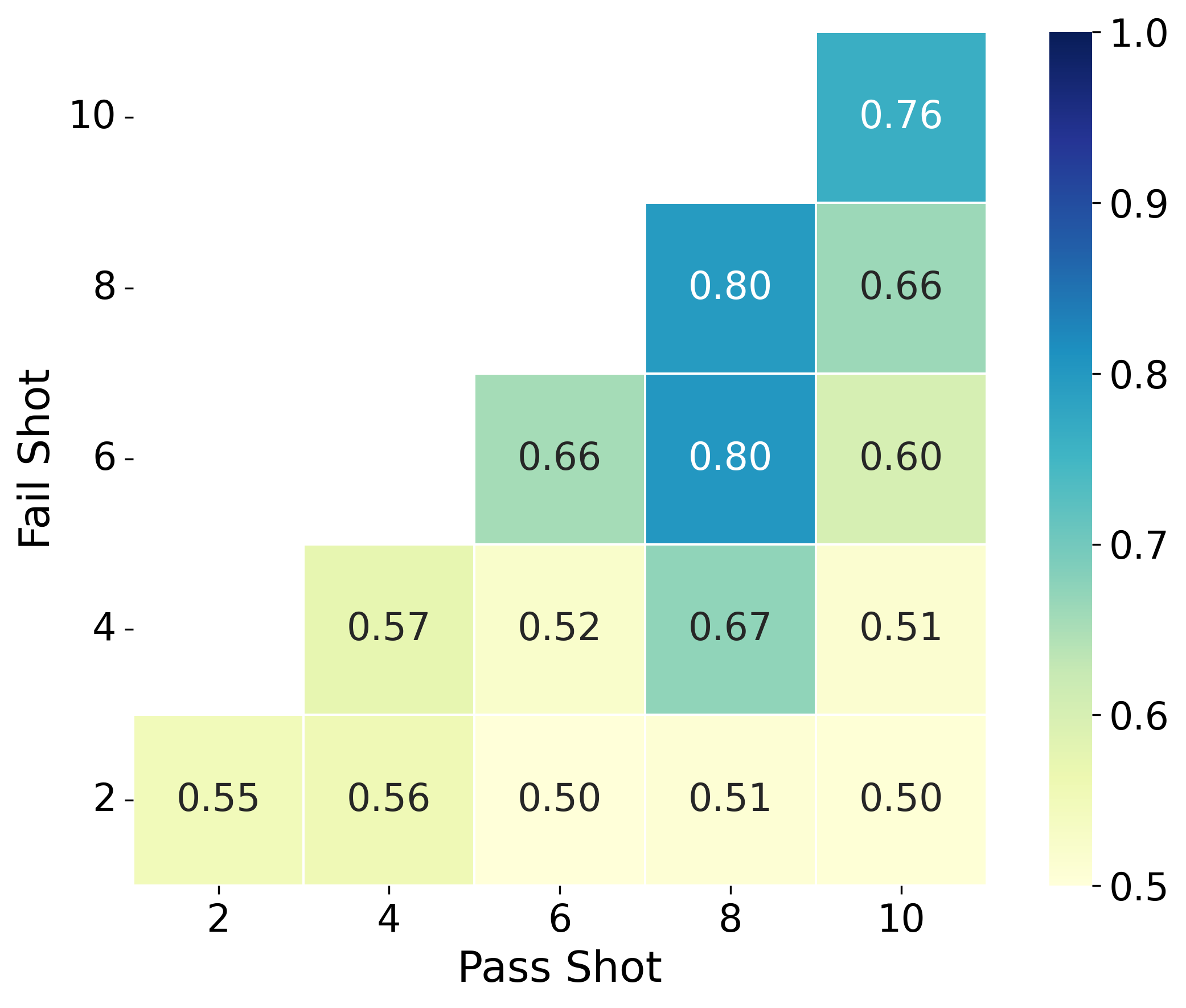}
        \caption{FS-Real}
        \label{fig:heatmap_imbalanced_fs_model}
    \end{subfigure}
    \hfill
    \begin{subfigure}[b]{0.33\linewidth}
        \centering
        \includegraphics[width=\linewidth]{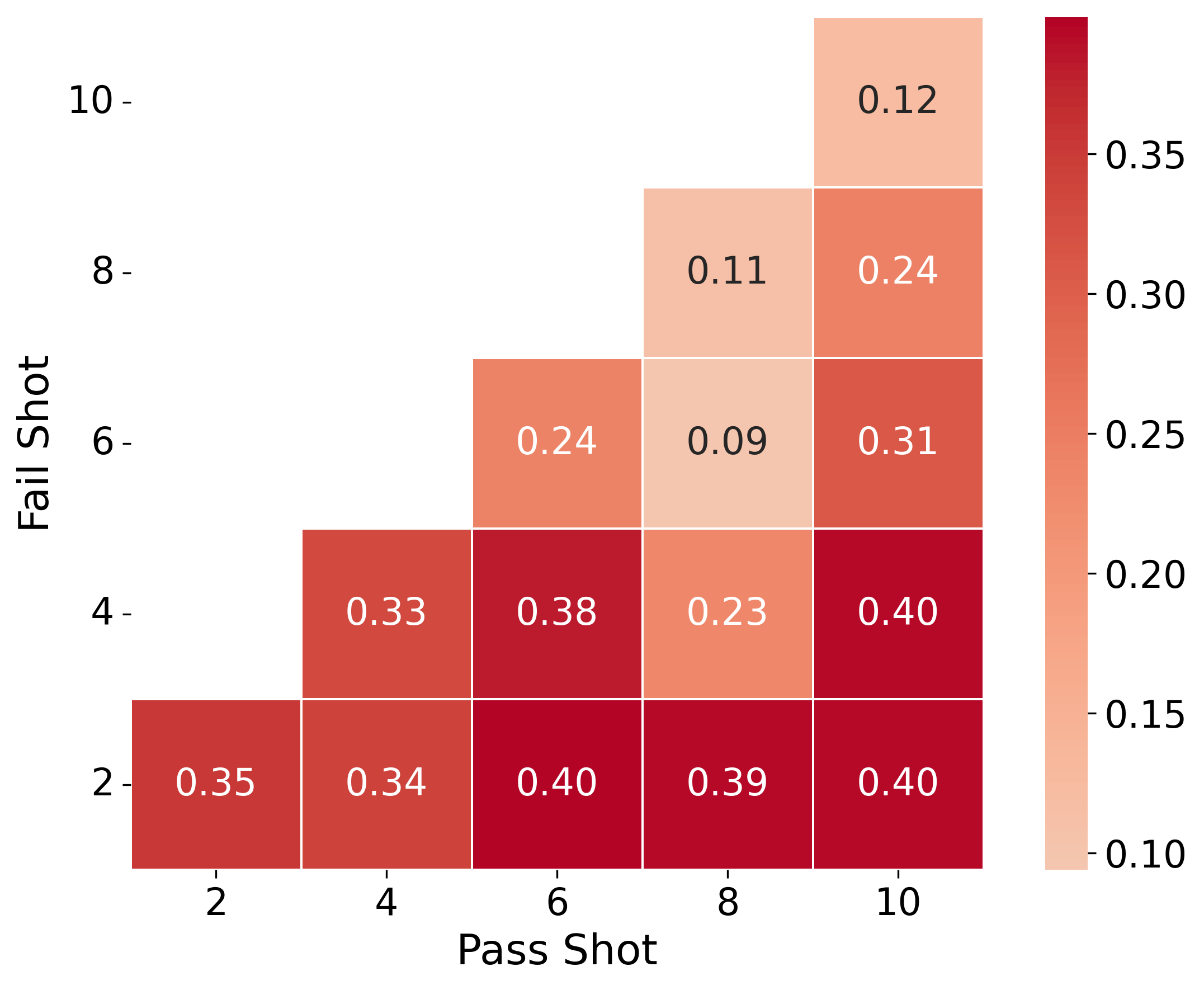}
        \caption{Performance difference}
        \label{fig:heatmap_imbalanced_diff_two_model}
    \end{subfigure}
    \caption{Balanced accuracy comparison across balanced and imbalanced pass and fail shot combinations. (a) SDG maintains stable performance (0.89--0.91) regardless of class imbalance. (b) FS-Real shows substantial degradation with increased imbalance. (c) Performance difference (SDG minus FS-Real) demonstrates consistent superiority of synthetic data approach. Please note: numbers of pass and fail images specified in heatmaps are randomly selected for training the few-shot baseline (FS-Real), with remaining images reserved for evaluation of both models.}
    \label{fig:heatmap_imbalanced_comparison}
\end{figure*}

Figure~\ref{fig:heatmap_imbalanced_comparison} presents balanced accuracy across imbalanced configurations. The SDG model maintains accuracy ranging from 0.89 to 0.91 across all tested combinations, with minimal variance independent of class imbalance ratio. The FS-Real model exhibits accuracy degradation correlated with increasing imbalance, ranging from 0.50 to 0.80 across configurations. For the 2 fail shot scenario, FS-Real accuracy remains near 0.50 despite pass shots increasing from 2 to 10, indicating training instability with limited fail examples. The performance difference between methods (Figure~\ref{fig:heatmap_imbalanced_diff_two_model}) ranges from 0.12 to 0.40 across configurations, with the maximum gap occurring at the most extreme imbalance (10 pass shots, 2 fail shots). This observation aligns with the real evaluation dataset distribution (pass : fail = 275 : 25 $\approx$ 11 : 1), indicating production-relevant performance characteristics.

\section{Discussion}
\label{sec:discussion}

\subsection{Labeling process in the real-world}
\label{sec:realworld-label-process}

Real-world data labeling is time-consuming and labor-intensive, which often lead to significant delays in factory-floor decision-making. A typical industrial workflow for data labeling proceeds as follows. Industry standards must first be established to ensure consistent production of conforming parts. Human experts then rely on these predefined standards to annotate the collected manufacturing data. In the case of the bracket studied in this research, parts must first be produced and measured by the vision system. Subsequently, human experts review the images, draw bounding boxes, and assign pass/fail labels. The time and effort required for this process are substantial. Moreover, because defective parts are relatively rare in real production environments, collecting sufficient samples across all quality classes can take considerable time.

Modern manufacturing increasingly demands responsive and flexible production to meet just-in-time requirements. When new parts are introduced, the entire data collection and labeling workflow must be repeated, resulting in additional delays. Both data collection procedures and the associated industry standards and annotation criteria must adapt to these dynamic production scenarios. The hybrid SDG pipeline with domain randomization proposed in this study enables annotation-free and scalable inspection systems that can be rapidly deployed in production.

\subsection{Model performance and error characterization}
\label{sec:error-characterization}

\begin{figure}[h]
    \centering
    \begin{subfigure}[b]{1\linewidth}
        \centering
        \includegraphics[width=\linewidth]{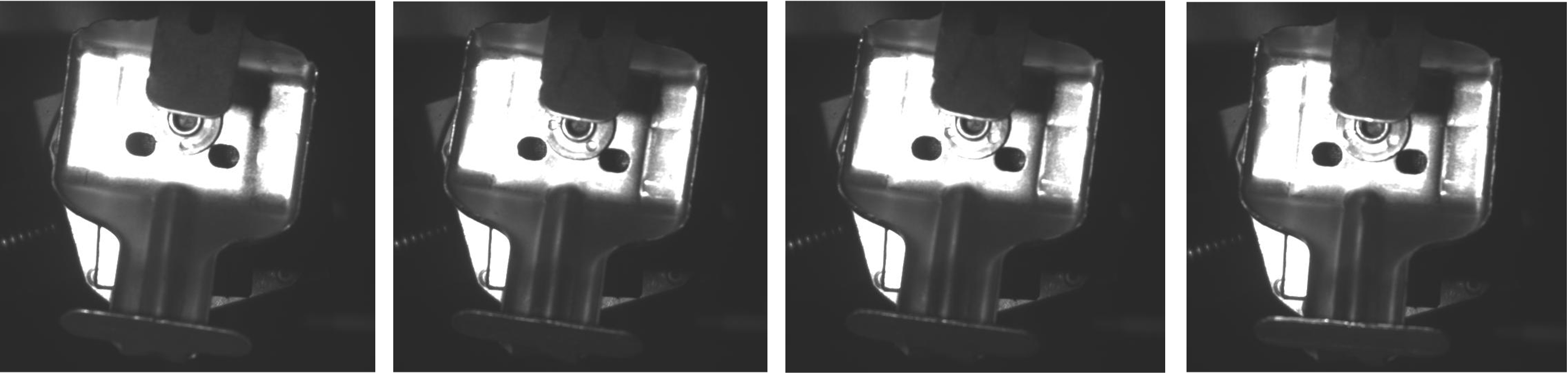}
        \caption{Type I errors from SDG model (true label = pass, prediction = fail).}
        \label{fig:sdgmodel_TruePass_PredictFail}
    \end{subfigure}
    \hfill
    \begin{subfigure}[b]{1\linewidth}
        \includegraphics[width=\linewidth]{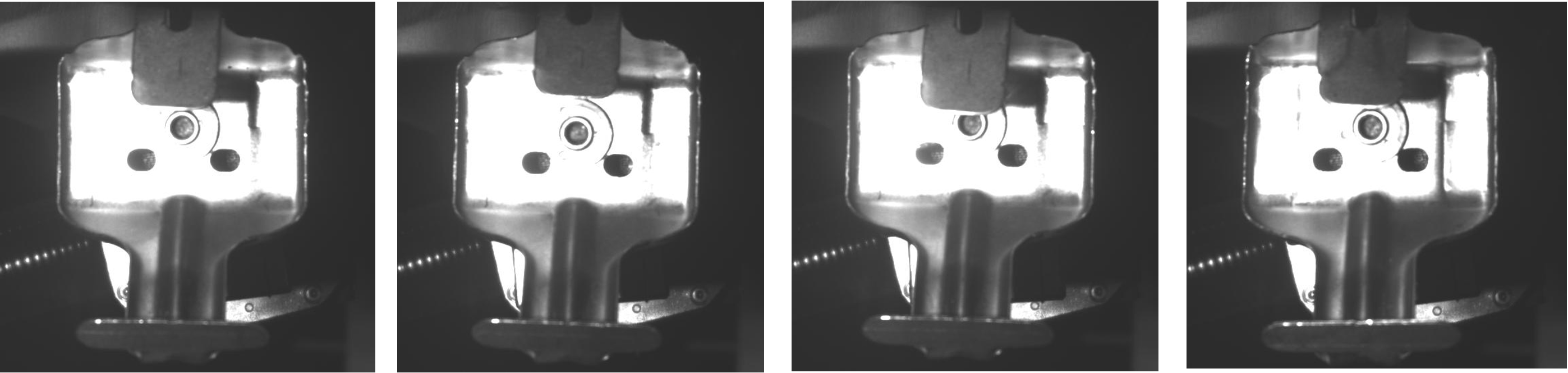}
        \caption{Type II errors from SDG model (true label = fail, prediction = pass).}
        \label{fig:sdgmodel_TrueFail_PredictPass}
    \end{subfigure}
    \caption{Examples of misclassified samples.}
    \label{fig:mobilenet-missclassification}
\end{figure}

The SDG model achieves 96.0\% accuracy and 90.1\% balanced accuracy on 300 real production images, with 288 correct classifications and 12 misclassifications (8 Type I errors, 4 Type II errors) as shown in Figure~\ref{fig:cm_sdg_model}. The combination of high accuracy and high balanced accuracy demonstrates both strong overall performance and robustness under the 11:1 class imbalance present in real data. Figure~\ref{fig:mobilenet-missclassification} shows representative misclassified samples categorized by error type. Type I errors (Figure~\ref{fig:sdgmodel_TruePass_PredictFail}) correspond to parts near the decision boundary where bending angles approach specification limits, making classification challenging even for human experts. Type II errors (Figure~\ref{fig:sdgmodel_TrueFail_PredictPass}) occur under extreme specular reflection from metallic surfaces under high-intensity lighting. While the SDG pipeline incorporates domain randomization for part geometry, lighting power, and surface roughness, it inadequately captures high-intensity specular reflection events observed in real inspection environments. Expanding randomization to include variable light source angles and increased brightness perturbation ranges may reduce Type II error rates.

\subsection{Extensibility to fine-scale quality assessment}
\label{sec:multiclass-classification}

Real-world manufactured parts demonstrate various geometric deformations along a continuous spectrum. In this work, discrete bending angles are used to create different part geometries, which are subsequently classified into binary pass and fail categories according to industry quality standards. However, the proposed SDG methodology is extensible to multi-class classification or even regression scenarios. Continuous part geometry variations can be realized by adjusting the parametric bending angle values in the CAD model, enabling fine-scale quality assessment beyond binary decisions. Multi-class classification or regression would provide manufacturers with detailed severity information for non-conforming parts, allowing for more adaptive and intelligent process control decisions. Since large-scale data collection in manufacturing requires domain expertise and is both labor-intensive and time-consuming, the ability to generate synthetic, labeled training data at scale presents opportunities for implementing intelligent quality control systems in data-scarce production environments.

\subsection{Sim-to-real generalization}
\label{sec:sim2real-transfer}

The performance comparison between SDG and FS-Real (Sections~\ref{sec:balanced-few-shot-baseline} and~\ref{sec:imbalanced-few-shot-baseline}) demonstrates that SDG consistently outperforms FS-Real across balanced and imbalanced few-shot scenarios. The SDG model, trained on synthetic images, achieves 90.1\% balanced accuracy when evaluated on real inspection images, indicating successful sim-to-real transfer. Domain randomization for part geometry, lighting conditions, and surface properties enables the model to learn features that generalize to real manufacturing data. However, the misclassification analysis reveals two limitations: (1) boundary cases where geometric features are marginally outside specification limits, and (2) lighting conditions (specular reflection) that fall outside the randomization parameter ranges used during SDG. These observations suggest that iterative refinement of the SDG method and quality inspection model based on systematic failure mode analysis can further narrow the sim-to-real gap. This also opens up future research direction of investigating active learning approaches where misclassified real samples inform updates to the domain randomization and adaptation strategy.

\subsection{Limitations and future work}
\label{sec:limits-and-future-work}

While the proposed hybrid SDG approach demonstrates strong zero-shot performance for industrial inspection, several limitations mentioned below provides opportunities for future research directions. First, this work is validated on a single part geometry and inspection configuration. Generalization to other part types, surface textures, and sensor setups requires additional investigation. Second, the domain randomization strategy assumes environmental conditions remain within the parameter ranges explored during SDG, and model performance may degrade under conditions significantly outside these ranges. Third, while synthetic dataset creation is substantially less labor-intensive than real data annotation, it requires computational resources and simulation expertise during setup. Finally, the comparison is primarily with few-shot baselines. Comprehensive evaluation against other domain adaptation techniques (e.g., domain-adversarial networks, self-supervised learning) would provide a broader context. 

Future work could address these gaps by: (1) bridging the sim-to-real gap through domain adaptation with mixed synthetic and real data, (2) validating the SDG pipeline on diverse parts and manufacturing processes to assess generalization, (3) developing adaptive domain randomization strategies incorporating real-world feedback to extend synthetic diversity, (4) integrating learned generative enhancements to model challenging phenomena such as specular reflections, and (5) extending the approach to multi-class and fine-grained defect assessment.

\section{Conclusions}
\label{sec:conclusions}

This paper presents a hybrid SDG framework combining simulation-based rendering, domain randomization, and real background compositing to enable zero-shot learning for computer vision-based industrial part inspection. The SDG pipeline generates 12,960 labeled images in one hour by varying part geometry, lighting, and surface properties, then compositing synthetic parts onto real image backgrounds. The SDG pipeline generates 12,960 labeled images in one hour by varying part geometry, lighting, and surface properties, and compositing synthetic parts onto real image backgrounds. Trained exclusively on synthetic data, the two-stage quality inspection architecture achieves part detection of mAP@0.5 of 0.995, classification accuracy of 96.0\%, and balanced accuracy of 90.1\% on real part images. Comparative evaluation shows the SDG model maintains 90–91\% balanced accuracy under severe imbalance (11:1 pass/fail ratio), achieving 23.3\% improvement over few-shot real-data baselines (90.0\% vs. 66.7\% in balanced scenarios. 89–91\% vs. 50\% under imbalanced scenarios). Error analysis reveals that failures concentrate in geometric boundary cases and underrepresented specular reflection conditions. Overall, this research demonstrates that the proposed method enables zero-shot learning for industrial part inspection without manual annotation or labeled real data, thereby significantly improving the efficiency and scalability of deploying visual inspection systems for new products. The proposed method is readily extensible to other manufacturing visual inspection tasks.

\section*{Acknowledgments}
\label{sec:acknowledgements}
The authors acknowledge the financial support of General Motors (GM) and the technical assistance provided by Norman Leo from GM during the Blender setup of this study.

\bibliography{reference}
\bibliographystyle{elsarticle-num}

\end{document}